\documentclass[letterpaper]{article} 
\usepackage[preprint]{aaai2027}  
\usepackage[hyphens]{url}  
\usepackage{graphicx} 
\usepackage{natbib}  
\usepackage{caption} 
\usepackage{algorithm}
\usepackage{algorithmic}
\usepackage{amsfonts, amsmath}

\usepackage{booktabs}
\usepackage{xcolor}
\usepackage{subcaption}

\usepackage{booktabs}
\usepackage[table]{xcolor}

\definecolor{lightblue}{RGB}{220,230,250}
\definecolor{lightgreen}{RGB}{220,245,220}
\definecolor{lightgray}{RGB}{245,245,245}
\usepackage{newfloat}
\usepackage{listings}
\DeclareCaptionStyle{ruled}{labelfont=normalfont,labelsep=colon,strut=off} 
\floatstyle{ruled}
\newfloat{listing}{tb}{lst}{}
\floatname{listing}{Listing}

\usepackage{booktabs}

\title{DCRA: Diffusion-Conditioned Representation Alignment for Robust Time-Series Learning}
\author{
    Wenrui Xu, Anas Enanaa, Keshab K. Parhi
}
\affiliations{
    University of Minnesota, Twin Cities\\
    \{xu000424, enana003, parhi\}@umn.edu
}

\begin{document}

\maketitle

\begin{abstract}
Learning robust representations for time-series signals under noise and distribution shifts remains challenging, especially in clinical applications such as electroencephalogram (EEG) and electrocardiogram (ECG) analysis. We propose \textit{Diffusion-Conditioned Representation Alignment} (DCRA), a training framework that repurposes the forward diffusion process as a structured corruption scheduler for representation learning.
Different from conventional augmentation and consistency-based methods that rely on independently sampled perturbations, DCRA introduces a structured corruption trajectory via the diffusion forward process, which enables continuous and controlled representation evolution across noise levels.
We introduce a feature-level consistency objective that aligns representations across noise levels while preserving class-discriminative structure. This mechanism promotes structure-preserving consistency, which enables smooth and semantically coherent feature trajectories in latent space. 
The proposed framework is encoder-agnostic and can be integrated with state space models and Transformer architectures. The seizure detection experiments on the CHB-MIT EEG dataset show that DCRA consistently improves performance under multiple noise conditions and achieves higher sensitivity at low false-positive rates. Analysis reveals that DCRA produces more balanced and structured representations compared to baseline and diffusion-only models. These findings highlight the benefit of combining structured corruption with representation alignment for robust time-series learning.
\end{abstract}


\section{Introduction}

Robust representation learning for long time-series data remains a persistent challenge under noise and distribution shifts. In real-world and clinical scenarios, time-series signals, such as electroencephalogram (EEG) and electrocardiogram (ECG) recordings, are commonly corrupted by noise, artifacts, and distribution shifts. Reliable modeling and generalization for time-series data analysis struggle with these corruptions. This issue is critical in EEG analysis, which is contaminated by muscle artifacts, ocular movements, baseline drift~\cite{islam2016methods}, electrode displacement, and recording variability across institutions~\cite{obeid2016temple}. These confounding factors may influence signal morphology and temporal dynamics, which leads to unstable feature representations. For clinical applications such as seizure detection, the robustness of the model is an important standard to evaluate the model's performance. The model must maintain high sensitivity under strict false-positive rate (FPR) constraints, even at low signal-to-noise ratios (SNRs). However, modern sequence encoders, such as convolutional neural networks, Transformers, and State Space Models (SSMs), often exhibit significant performance degradation when exposed to low-SNR conditions~\cite{zhang2021survey} or distribution shifts.

Noise augmentation has been widely used to improve the generalization and robustness of models~\cite{Hendrycks_2021_ICCV}, but conventional methods typically treat noise as independent perturbations~\cite{cubuk2020randaugment,zhang2017mixup} without enforcing representation invariance. As a result, an encoder will overfit to specific corruption patterns rather than preserve consistent features across noise levels. The recent advancements in physics-informed diffusion begin to address the lack of principal mechanisms to model structured corruption trajectories. 
Diffusion Probabilistic Models (DPMs)~\cite{sohldickstein2015deepunsupervisedlearningusing} provide a multi-scale corruption process through the forward noise dynamics. Even though diffusion models are primarily developed for generative modeling, their forward process defines a well-structured family of progressive signal degradation across different noise scales. This observation motivates a significant question:

\textit{Can we repurpose the diffusion forward process as a corruption scheduler to learn noise-robust representations for discriminative time-series tasks?}

In this paper, we propose a Diffusion-Conditioned Representation Alignment (DCRA), an encoder-agnostic training framework that leverages diffusion-based corruption with feature-level consistency. Unlike prior diffusion-based approaches that rely on reverse denoising or generative reconstruction, we leverage the forward diffusion process as a structured corruption scheduler, which aims to feed the encoder multiple corruptions of the same input signal. Subsequently, a representation alignment objective is introduced to encourage clean and corrupted signals to produce noise-robust, semantically consistent, and class-discriminative latent features. 


We evaluate the proposed framework on the widely used clinical EEG benchmark CHB-MIT~\cite{PhysioNet-chbmit-1.0.0} dataset using ROC-AUC (AUC), PR-AUC, and sensitivity under fixed FPR.
Experimental results show improved robustness across multiple noise levels and enhanced sensitivity at low FPR rates. These findings indicate that structured corruption modeling integrated with latent feature alignment provides an effective strategy for robust representation learning. The main contributions of this paper are summarized as follows:
\begin{itemize}
    \item We introduce a novel training framework that makes use of the forward diffusion process as a structured multi-scale corruption scheduler for time-series representation learning.
    \item We propose a feature-level consistency mechanism that enforces consistency between clean and corrupted signals in the latent space.
    \item We demonstrate improved robustness and fixed-FPR sensitivity on the clinically distinct EEG benchmark under multiple noise levels. 
\end{itemize}

The paper is organized as follows. Section II reviews the related works in representation learning, diffusion models for structured corruption, and seizure detection. Section III introduces the basics of State Space Models and Diffusion Probabilistic Models. Section IV demonstrates the proposed training framework. Sections V and VI discuss the experimental setup and results.
\section{Related Work}\label{sec:related_work}
\subsection{Representation Learning via Data Augmentation and Consistency Regularization}
Data augmentation techniques have been widely used to improve the generalization capability and robustness of deep learning models by injecting perturbations during training. In the field of computer vision and image processing, automated augmentation strategies, such as AutoAugment~\cite{cubuk1805autoaugment} and RandAugment~\cite{cubuk2020randaugment}, systematically search for data augmentation policies that enhance the performance of models. Mixup~\cite{zhang2017mixup} regularization improves robustness to adversarial examples and stabilizes training by interpolating training samples. Advancing beyond these empirical augmentation techniques, several works introduce consistency constraints to enforce stability in the representation domain. Contrastive learning frameworks, including SimCLR~\cite{chen2020simple} and MoCo~\cite{he2020momentum}, learn invariant features by maximizing agreements between representations of different augmented views. Similarly, Semi-supervised methods, such as Mean Teacher~\cite{tarvainen2017mean} and FixMatch~\cite{sohn2020fixmatch}, introduce consistency regularization under stochastic perturbations. 
These approaches are closely related to self-training paradigms, where models are encouraged to produce stable predictions across perturbed inputs or pseudo-labels~\cite{xie2020self}. Empirical studies have shown that the effectiveness of consistency-based methods is dependent on the design of perturbations and evaluation protocols~\cite{oliver2018realistic}.
However, most of these approaches treat perturbations as independent transformations rather than modeling corruptions as structured trajectories. In contrast, our proposed framework adopts a diffusion-based corruption process that generates degraded signals with different noise levels and enforces cross-noise representation space alignment along the structured path. 
\subsection{Diffusion Model for Structured Corruption}
Diffusion models, such as Denoising Diffusion Probabilistic Models (DDPM)~\cite{ho2020denoisingdiffusionprobabilisticmodels} and score-based generative modeling~\cite{song2020score}. 
Diffusion has primarily been explored for generative modeling, but recent studies, including Diffusion-based Representation Learning (DRL)~\cite{mittal2023diffusion}, have investigated its potential for representation learning and robustness. Most prior work relies on the reverse denoising process or generative reconstruction. In contrast, the DCRA framework takes a different perspective, where we exploit the forward process as a structured noise scheduler and directly incorporate it into the supervised learning objective. The proposed framework converts diffusion from a generative modeling tool to a training strategy that constrains representations along a structured diffusion trajectory, which provides control over feature evolution across noise levels. 

\subsection{Seizure Detection}


Robustness of classification for physiological signals, including EEG and ECG, is critical for clinical applications such as seizure detection and arrhythmia detection. Deep learning models, including CNNs, recurrent networks, and Transformers, have achieved strong performance in EEG seizure detection~\cite{lawhern2018eegnet,bashivan2015learning,song2022eeg}. While existing studies often employ heuristic augmentation or artifact removal to improve robustness, they rarely enforce representation consistency across multiple corruption levels. Our framework addresses this limitation through structured diffusion conditioning and cross-noise representation regularization.

\section{Background}\label{sec:background}

In this section, we review the theoretical foundations of key components underlying our proposed framework, including State Space Models and Diffusion Probabilistic Models
which constitute the conceptual basis of our proposed architecture.

\subsection{State Space Models and Mamba}\label{subsec:SSMs}
State Space Models (SSMs) model sequential dynamics through a latent state that evolves according to a linear dynamical system~\cite{gu2020hipporecurrentmemoryoptimal}. Given an input $x(t)$, the hidden state $h(t)$ and output $y(t)$ are defined as

\begin{equation}
    \dot{h}(t) = \mathbf{A}h(t) + \mathbf{B}x(t), \quad y(t) = \mathbf{C}h(t) \nonumber
\label{eq:continuous SSMs}
\end{equation}
where $\mathbf A$, $\mathbf B$, and $\mathbf C$ denote the state transition, input, and output matrices, respectively.
In practice, in order to process discrete, high-frequency neural sequences such as iEEG~\cite{puah2025eegdmeegrepresentationlearning} and ECG~\cite{qiang2024ecgmambaefficientecgclassification} signals, the continuous system is discretized using the Zero-Order Hold (ZOH) rule.


Mamba~\cite{gu2024mambalineartimesequencemodeling} advances standard SSMs by introducing input-dependent state updates through a selective scan mechanism. Compared with Transformer architectures, Mamba achieves linear computational complexity with respect to sequence length while maintaining strong long-range modeling capability, which makes it suitable for long, high-frequency time-series data analysis.

\subsection{Diffusion Probabilistic Models and Diffusion Models}\label{subsec:DPMs}
Diffusion probabilistic models (DPMs) progressively corrupt clean data by adding Gaussian noise over multiple timesteps~\cite{sohldickstein2015deepunsupervisedlearningusing}.
Following the variance-preserving Markov chain, the intermediate latent $x_t$ at any timestep $t \in [0, T]$ is sampled in closed-form expression:
\begin{equation}
    x_t = \sqrt{\bar{\alpha}_t} x_0 + \sqrt{1 - \bar{\alpha}_t} \epsilon, \quad \epsilon \sim \mathcal{N}(0, \mathbf{I}),
\end{equation}
where $\bar{\alpha}_t = \prod_{i=1}^t \alpha_i$ is the cumulative product of $\{\alpha_i\}s$ and $\alpha_t = 1 - \beta_t$ denotes the noise schedule. The forward process variance term \{$\beta_t\}_{t=1}^T$ controls the magnitude of injected noises. 
Different from generative diffusion models that learn reverse denoising, DCRA uses only the forward diffusion process as a structured corruption scheduler to encourage the encoder to learn noise-robust representations that remain stable across varying signal-to-noise ratios (SNRs).

\begin{figure*}[htbp]
    \centering
    \includegraphics[width=\linewidth]{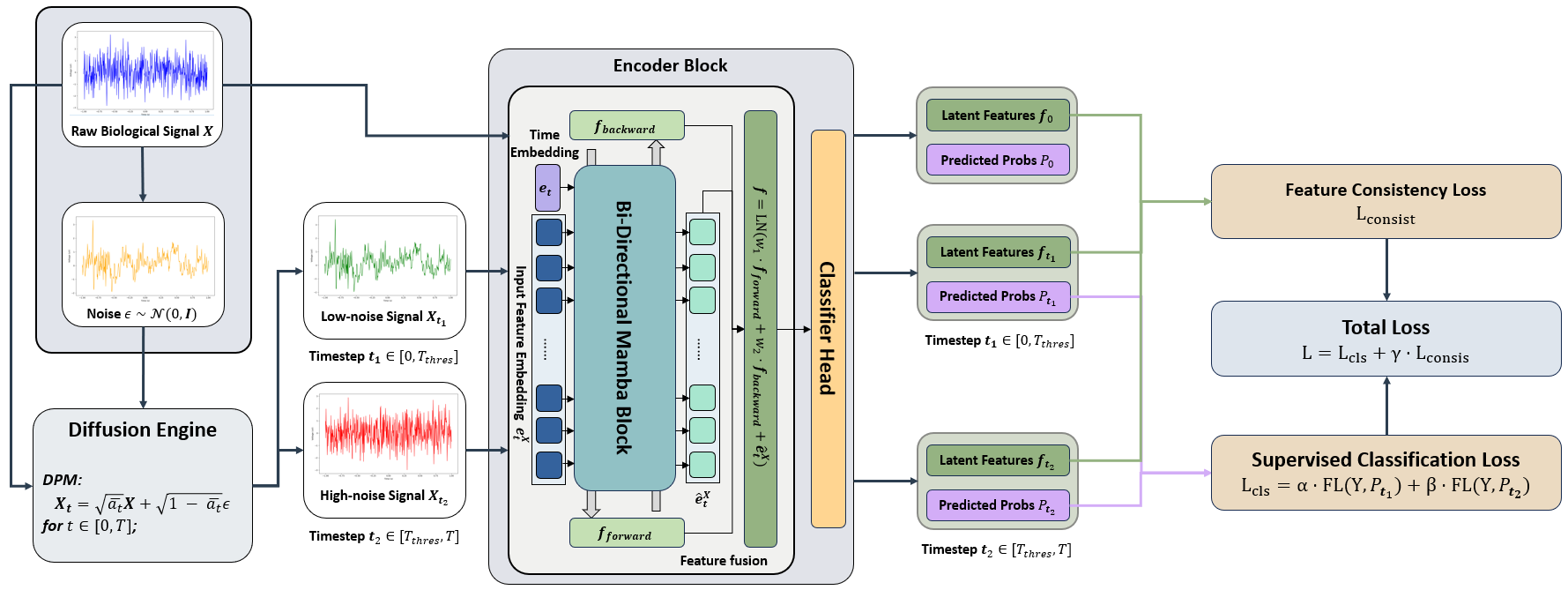}
    \caption{\small{Overview of the proposed diffusion-conditioned representation alignment (DCRA) framework for time-series analysis. The model integrates a dual-timestep diffusion process with a shared encoder to learn noise-robust representations across multiple corruption levels. A diffusion-conditioned (DC) encoder (e.g., Bi-directional Mamba) processes clean and noisy signals sequentially with timestep embeddings to capture long-range temporal dependencies. The framework is trained with a hybrid objective that combines supervised classification loss and cross-noise consistency regularization, which enforces both noise-robust representations and class-discriminative latent structure.}}
    \label{fig:framework}
\end{figure*}

\section{Methodology}\label{sec:method}

In this section, we propose the diffusion-conditioned Representation Alignment (DCRA) framework, illustrated in Fig.~\ref{fig:framework}. Given an input signal, two corrupted views are generated through forward diffusion at different noise levels and processed by a shared diffusion-conditioned encoder. The model is optimized using supervised classification together with cross-noise consistency regularization, encouraging robust and class-discriminative representations across the diffusion trajectory.

\subsection{Dual-timestep Diffusion Engine}


For an input signal $X\in\mathbb{R}^{C\times L}$ with
$C$ channels and $L$ input length, two corrupted views are generated using the forward diffusion process $X_t=\sqrt{\bar{\alpha}_t}X+\sqrt{1-\bar{\alpha}_t}\epsilon$, where $\epsilon\sim\mathcal N(0,I)$ and $\bar{\alpha}_t$ is determined by the predefined noise schedule.
This formulation provides a structured corruption process indexed by timestep $t$ rather than independent perturbations, which allows controlled generation of degraded signals at different noise levels. 
Then we introduce the Dual-Timestep Sampling strategy in order to enforce robustness across noise intensities. For each sample $X$, two timesteps are sampled:
$t_1\in [0, T_{thres}] \text{, }t_2 \in [T_{thres}, T]$,
where $t_1$ and $t_2$ correspond to low-noise and high-noise levels, respectively. The threshold $T_{thres}$ is dynamically selected to ensure a clear separation between low- and high-noise signals.


\subsection{Diffusion-conditioned Encoder}

The proposed framework is not restricted to a specific encoder architecture. Biological time-series signals, such as EEG signals, often exhibit long-range and non-causal temporal dependencies (e.g., seizure onsets). In order to better capture these patterns, we employ a bi-directional state space encoder based on the Mamba architecture. The clean and corrupted signals are then fed into the encoder, respectively.

In order to make the encoder aware of the corruption intensity, the model is built on the diffusion timestep. For a given corrupted signal $X_t$, the timestep $t$ is first mapped to a continuous timestep scalar $\hat{t}\in [0, 1]$
: 
 $\hat{t} = \frac{t}{T - 1} \in [0, 1]$ for $t\in \{0, 1, 2,\dots, T-1\}$.
The normalized timestep $\hat{t}$ is then projected into a learnable latent embedding $e_t$ via a Multilayer Perceptron (MLP): $e_t = \text{MLP}(\hat{t}) \in \mathbb{R}^{d}$, where $d$ is the feature dimension. Subsequently, the timestep embedding is injected into the feature embedding of the input signal $e^X_t \in \mathbb{R}^{L \times d}$ through additive conditioning: 
$$\hat{e}^X_t = e^X_t + \mathbf{1}_Le_t^\top,$$ 
where $\mathbf{1}_L$ represents a row vector with dimension $L$ and $\mathbf{1}_Le_t^\top$ represents the transposed feature embedding $\mathbf{1}_Le_t$.
This conditioning mechanism enables the encoder to adapt its internal dynamics according to the corruption intensities while preserving a robust representation learning capability across different noise levels.

The bi-directional Mamba block $\text{Mamba}(\cdot)$ is implemented as the encoder to model the temporal dependencies and non-causality of biological signals. The conditioned sequence $\hat{e}^X_t$ is processed in both temporal directions:
\begin{align}
    &f_{forward} = h_{fwd} = \text{Mamba($\hat{e}^X_t$)},\\& f_{backward} = h_{bwd} = \text{Flip}(\text{Mamba($\text{Flip}(\hat{e}^X_t$))}), \nonumber
\end{align}
where $\text{Flip}(\cdot)$ denotes the flipping operation. The resulting feature maps $f_t$ are integrated with the original input through a residual fusion followed by a layer normalization to stabilize the latent distribution: 
\begin{align}
    f_t = \text{LayerNorm}(\hat{e}^X_t + w_1 \cdot f_{forward} + w_2 \cdot f_{backward}), \nonumber
\end{align} 
where $w_1$ and $w_2$ denote feature fusion weights. We further introduce a set of learnable class prototypes to structure the latent space:
$\mathbf{C} = \{c_1, \dots, c_K\}, c_k \in \mathbb{R}^d, $
which serve as anchors for class-specific feature alignment.
Finally, the predicted probabilities for classes are generated via an MLP-based classifier head $\text{Classifier}(\cdot)$ based on the refined latent representations $h_t$:
$P_t = \text{Classifier}(f_t)$.

\subsection{Joint Optimization Objective}

The proposed framework is trained with a hybrid objective that combines supervised classification with cross-noise consistency regularization. For the supervised classification loss, we adopt the Focal loss $\text{FL}(\cdot)$ to address class imbalance, which is commonly observed in biomedical datasets. For given predictions from the low-noise and high-noise branches, the classification objective is formulated as:
\begin{equation}
    \mathcal{{L}}_{cls} = \alpha \cdot \text{FL}(Y, P_{t_1}) + \beta\cdot\text{FL}(Y, P_{t_2}),
\end{equation}
where $Y$ denotes the ground truth labels, and $\alpha$, $\beta$ control the contribution of the two noise levels.

To enforce the robust representation in latent space, we introduce a cross-noise consistency regularizer that aligns representations across different corruption levels. The regularization consists of two key components: anchor-based subspace consistency and prototype-based alignment loss, which operate on both projection and feature space. A prototype head $\text{Proto}(\cdot)$ is first applied on the latent representations $f_t$ from encoder block for projected feature $z_t = \text{Proto}(f_t)$,
where $\text{Proto}(\cdot)$ is a lightweight MLP used for representation stabilization. For the anchor-based subspace consistency term,  we treat the clean features as anchors, and align noisy features toward them, which helps enforce noise-robust representations:
\begin{equation}
    \mathcal{L}_{sub} = \mathbb{E}\left[w_{\mathrm{consist}_1} \| z_{t_1} - z_{0} \|_2^2 + w_{\mathrm{consist}_2} \| z_{t_2} - z_{0} \|_2^2 \right],
\end{equation}
where the weights $w_{\mathrm{consist}_1}, w_{\mathrm{consist}_2}$ decay with increasing noise levels, i.e., $w_{\mathrm{consist}_i} = 1 - \hat{t}_i$. 
Inspired by~\cite{Deng_2022, wen2016discriminative, snell2017prototypical}, the prototype-based alignment loss enforces class-discriminative structure by aligning features with learnable class prototypes $\mathbf{C}$, which is defined as:
\begin{align}
\mathcal{L}_{\mathrm{proto}} &= \mathbb{E}\Big[ \ \mathrm{CE}\!\left(Y, \frac{h_0 \mathbf{C}^\top}{\tau}\right)
+ w_{\mathrm{consist}_1}\,\mathrm{CE}\!\left(Y, \frac{h_{t_1} \mathbf{C}^\top}{\tau}\right) \nonumber \ \\& +  w_{\mathrm{consist}_2}\,\mathrm{CE}\!\left(Y, \frac{h_{t_2} \mathbf{C}^\top}{\tau}\right) \Big],
\end{align}
where $\mathrm{CE}(\cdot)$ denotes the cross-entropy loss, $h_t$ denotes the pooled features $h_t = \frac{\mathrm{Pool}(f_t)} {\|\mathrm{Pool}(f_t)\|_2}\in R^d$, and $\tau$ is a temperature parameter. A prototype regularization term is employed to ensure the  batch-wise  consistency of class centers:    $\mathcal{L}_{reg} = \frac{1}{K}\cdot\sum_{k} \| \bar{h}_{0}^{(k)} - c_k \|_2^2,$
where $\bar{h}_{0}^{(k)}$ denotes the mean clean feature of class $k$ within the batch. The overall cross-noise consistency regularization is defined as:
\begin{align}
    \mathcal{L}_{consis} = \lambda_1 \mathcal{L}_{sub} 
    + \lambda_2 \mathcal{L}_{proto} 
    + \lambda_3 \mathcal{L}_{reg}.
\end{align}
This consistency objective promotes the local smoothness of the encoder with respect to the corruption levels while preserving class-dependent geometry in the latent space. The subspace consistency term enforces smoothness along the diffusion-induced corruption trajectory by aligning representations across noise levels. In contrast, the prototype-based loss introduces class-aware constraints that maintain discriminative structure and prevent feature collapse. These components encourage representations to evolve coherently across noise levels while preserving class separation, leading to improved robustness under noise corruptions and distribution shifts.
The final training objective is defined as: 
\begin{equation}
    \mathcal{L} = \mathcal{L}_{cls} +\gamma \cdot \mathcal{L}_{consis},
\end{equation}
where $\gamma$ is adaptively adjusted based on the relative magnitude of classification and consistency losses using an exponential moving average (EMA) tool, which stabilizes training between two regularization terms.

\section{Experimental Setup}\label{sec:setup}

The DCRA framework is evaluated for detection of seizures on the CHB-MIT dataset~\cite{PhysioNet-chbmit-1.0.0} using a patient-specific protocol. More specifically, the models are trained and tested independently for each subject. We design a multi-fold cross-validation experiment to ensure the reliability and clinical significance of the evaluation. The dataset of each patient is partitioned into multiple folds based on seizure events, where each fold contains at least one seizure segment that is completely excluded from the training and validation sets. This pattern guarantees that the model is evaluated on unseen seizure patterns, which prevents information leakage and ensures a realistic assessment of generalization to new seizure events.
The raw EEG signals are first preprocessed into 18 standard channels, such as FP1-F7, F7-T7, T7-P7, etc. The continuous recordings are then segmented into samples using a 2-second sliding window without overlap. For each fold, z-score normalization is computed using the training set and then applied to the validation and test sets to avoid data leakage. This experimental setup allows a controlled comparison between baseline encoders and their diffusion-conditioned versions with and without feature consistency, which ensures that improvements in performance can be attributed to the proposed DCRA framework.

\begin{table*}[htbp]

\centering
\caption{Performance comparison across different encoders and training strategies.}
\label{tab:performance_t0}

\resizebox{0.8\linewidth}{!}{
\begin{tabular}{lccccc}
\toprule

\rowcolor{lightgray}
\textbf{Model} & \textbf{AUC} & \textbf{PR-AUC} & \textbf{F1} & \textbf{Sens@FPR=0.05 (\%)} & \textbf{Sens@FPR=0.10 (\%)} \\

\midrule


EEGNet~\cite{lawhern2018eegnet}
& 0.93 $\pm$ 0.09 
& \textbf{0.69 $\pm$ 0.23} 
& \textbf{0.67 $\pm$ 0.22} 
& 86.62 $\pm$ 14.30 
& 88.30 $\pm$ 13.47 \\

\midrule

\multicolumn{6}{c}{\cellcolor{lightblue}\textbf{Mamba-based Encoder}} \\

Mamba
& 0.96 $\pm$ 0.05 
& 0.62 $\pm$ 0.20 
& 0.58 $\pm$ 0.18 
& 88.00 $\pm$ 12.15 
& 91.13 $\pm$ 10.02 \\

DC-Mamba 
& \textbf{0.97 $\pm$ 0.04}
& \textbf{0.66 $\pm$ 0.24} 
& 0.62 $\pm$ 0.22 
& 89.99 $\pm$ 11.24 
& \textbf{92.73 $\pm$ 9.02} \\

\rowcolor{lightgreen}
DCRA-Mamba
& \textbf{0.97 $\pm$ 0.04}
& \textbf{0.66 $\pm$ 0.24} 
& \textbf{0.65 $\pm$ 0.22} 
& \textbf{90.04 $\pm$ 10.40}
& 92.48 $\pm$ 8.08 \\

\midrule

\multicolumn{6}{c}{\cellcolor{lightblue}\textbf{Transformer-based Encoder}} \\

Transformer
& 0.94 $\pm$ 0.06 
& 0.45 $\pm$ 0.23 
& 0.42 $\pm$ 0.20 
& 80.78 $\pm$ 17.05 
& 85.96 $\pm$ 13.71 \\

DC-Transformer
& \textbf{0.95 $\pm$ 0.07} 
& 0.52 $\pm$ 0.25 
& 0.45 $\pm$ 0.24 
& 82.90 $\pm$ 19.40 
& 87.28 $\pm$ 15.29 \\

\rowcolor{lightgreen}
DCRA-Transformer
& 0.93 $\pm$ 0.10 
& \textbf{0.54 $\pm$ 0.27} 
& \textbf{0.51 $\pm$ 0.27} 
& \textbf{83.57 $\pm$ 19.19} 
& \textbf{87.39 $\pm$ 16.35} \\

\bottomrule
\end{tabular}
}

\end{table*}

\section{Experimental Results \& Analysis}\label{sec:result}

The experimental results are averaged across all subjects to provide a comprehensive evaluation of the proposed DCRA framework. The timestep $T$ is set to $1000$, and the threshold $T_{thres}$ is set to $500$. We analyze the effectiveness of DCRA from three perspectives, including performance, robustness, and representation behavior in latent space. EEGNet is applied as a baseline. 

We first compare different training strategies, including a pure encoder baseline, a diffusion-conditioned encoder without consistency regularization (DC-encoder), and the full DCRA framework with structured feature consistency (DCRA-encoder). This comparison helps identify the contributions of both diffusion conditioning and representation alignment as shown in Table~\ref{tab:performance_t0}. Diffusion conditioning improves PR-AUC and sensitivity under FPR constraints, which demonstrates its effectiveness in enhancing robustness. The incorporation of feature consistency does not degrade the overall performance; it improves detection sensitivity in the low-FPR constraints, such as Sens@0.05 and Sens@0.10, while maintaining or even improving PR-AUC. Although improvements in global metrics such as AUC are modest, DCRA consistently improves sensitivity under strict FPR constraints, which is more relevant for clinical deployment scenarios.










\begin{figure*}[htbp]
    \centering
    \begin{subfigure}[b]{0.48\textwidth}
        \centering
        \includegraphics[width=\textwidth]{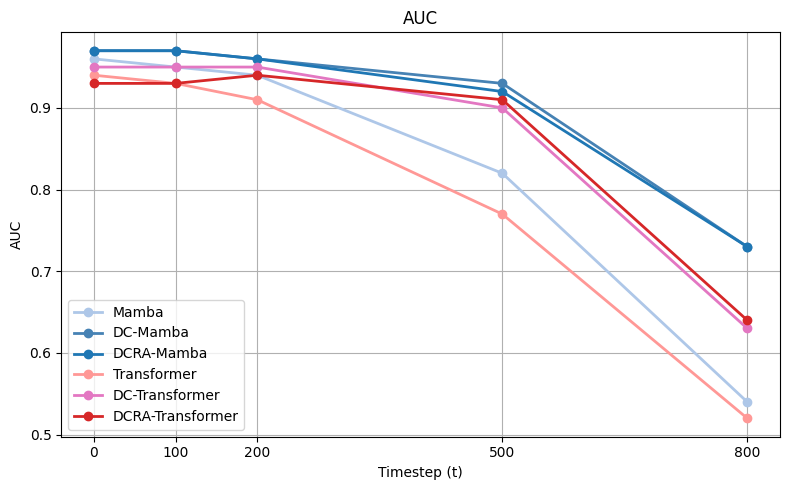}
        \caption{AUC Performance}
        \label{fig:auc}
    \end{subfigure}
    \hfill
    \begin{subfigure}[b]{0.48\textwidth}
        \centering
        \includegraphics[width=\textwidth]{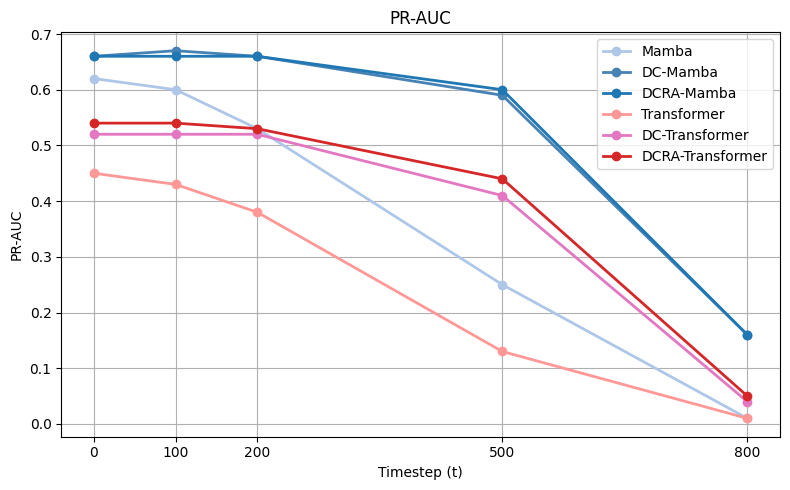}
        \caption{PR-AUC Performance}
        \label{fig:pr_auc}
    \end{subfigure}

    \vspace{1em} 

    \begin{subfigure}[b]{0.48\textwidth}
        \centering
        \includegraphics[width=\textwidth]{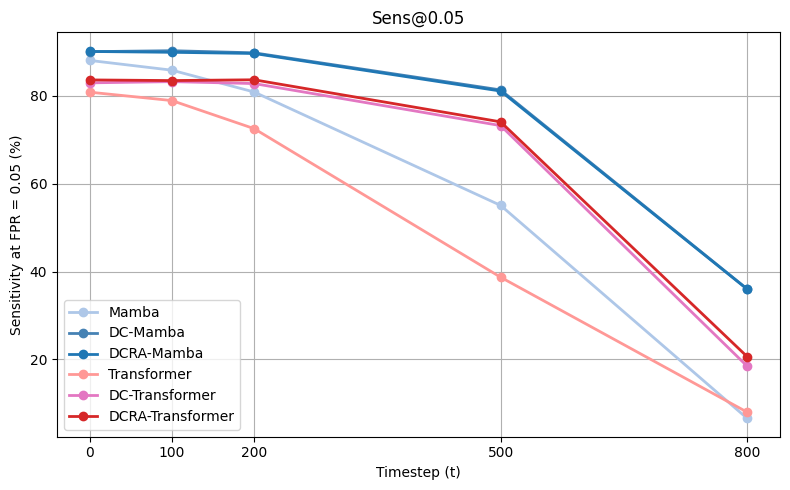}
        \caption{Sensitivity at FPR = 0.05}
        \label{fig:sens_005}
    \end{subfigure}
    \hfill
    \begin{subfigure}[b]{0.48\textwidth}
        \centering
        \includegraphics[width=\textwidth]{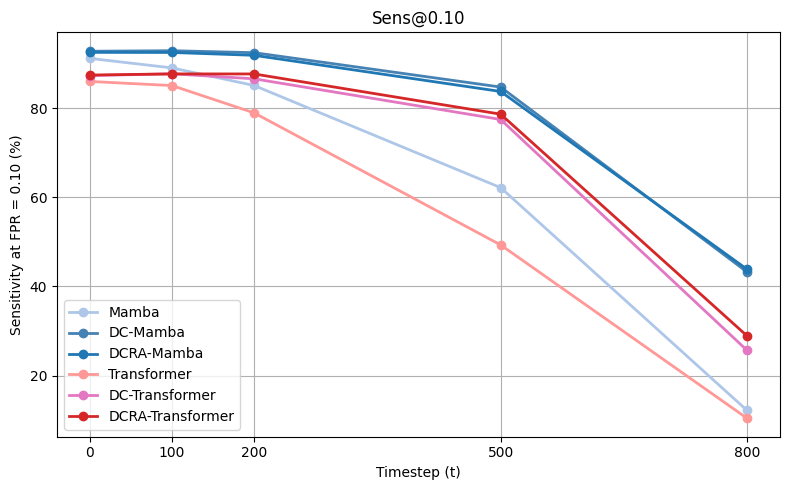}
        \caption{Sensitivity at FPR = 0.10}
        \label{fig:sens_010}
    \end{subfigure}

    \caption{\small{Performance of different encoders and strategies across various timesteps. The DCRA framework demonstrates stronger robustness to increased diffusion steps compared to baseline models.}}
    \label{fig:performance_cross_t}

\end{figure*}

We further evaluate model robustness under increasing noise levels by measuring performance across different diffusion timesteps, where SNR values keep decreasing. This analysis allows us to identify how well each method preserves discriminative capability as the corruption level increases. We interpret robustness as reflected by performance degradation curves across multiple noise levels.
As shown in Fig.~\ref{fig:performance_cross_t}, all encoders show degradation in performance as the noise level increases. However, DC and DCRA-encoders demonstrate improved robustness compared to the pure encoders. The DCRA-encoder maintains higher AUC and PR-AUC across intermediate-to-high noise conditions, which indicates improved robustness in preserving global discriminative structure in latent space. The main advantage of the proposed framework is reflected in low-FPR situations. As illustrated in Fig.~\ref{fig:sens_005} and Fig.~\ref{fig:sens_010}, the DCRA framework achieves higher sensitivity at low FPR, especially for noise levels $t \leq 500$ ($\text{SNR} \geq -20$ dB). While performance degrades under extreme corruption ($t=800$, $\text{SNR} \leq -35$ dB), the DCRA framework exhibits slower degradation trends across all encoders, which indicates the improved stability of learned representations across corruptions. As shown in Fig.~\ref{fig:pr_auc}, the feature consistency mechanism further improves robustness by preserving class-discriminative structure across corruption levels.

\begin{figure*}[htbp]
    \centering
    \begin{subfigure}[b]{\textwidth}
        \centering
        \includegraphics[width=\textwidth]{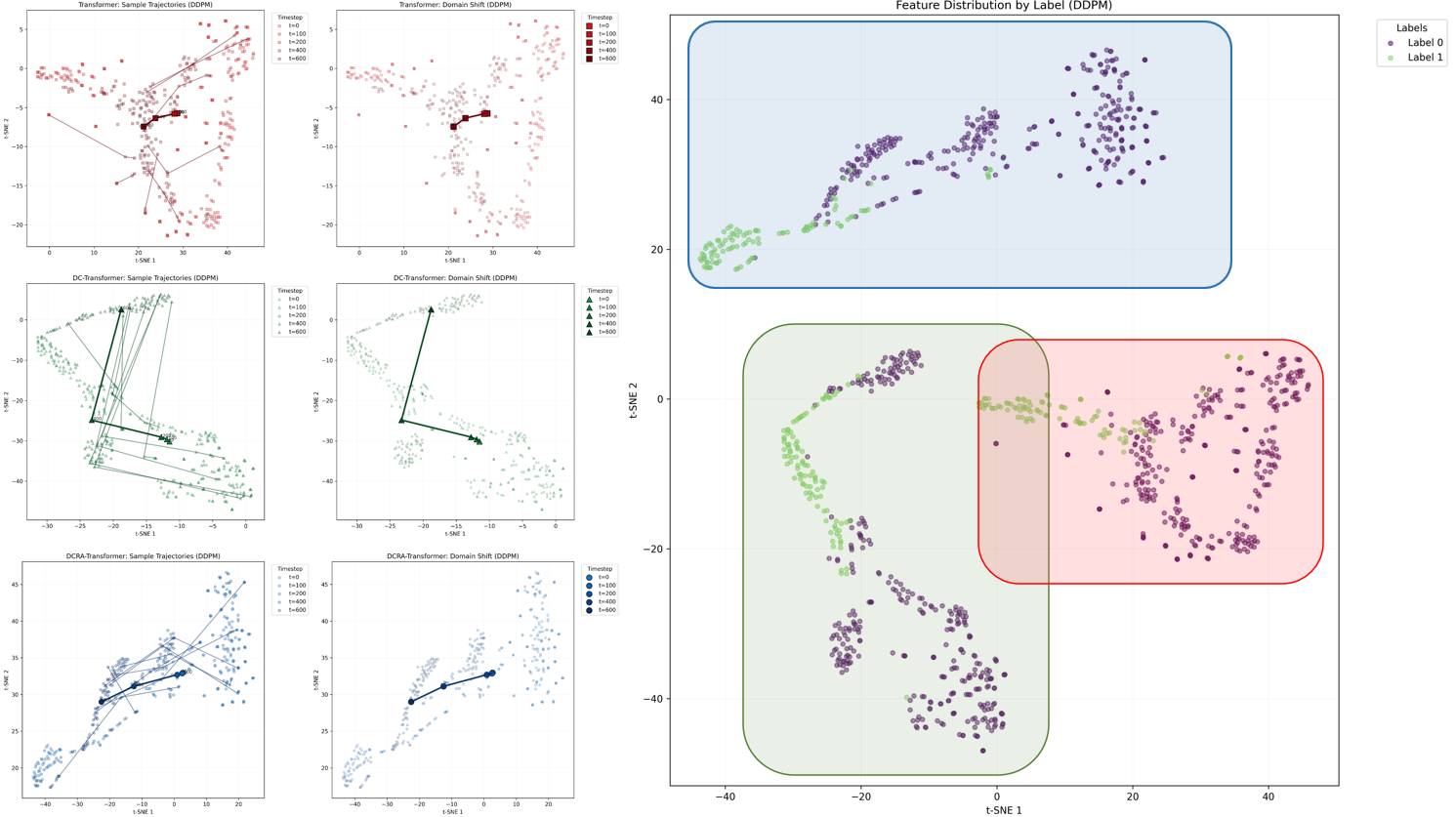}
        \caption{Feature Consistency Analysis of Transformer-based Models under Diffusion Noise}
        \label{fig:transformer_consistency}
    \end{subfigure}
    \hfill
    \begin{subfigure}[b]{\textwidth}
        \centering
        \includegraphics[width=\textwidth]{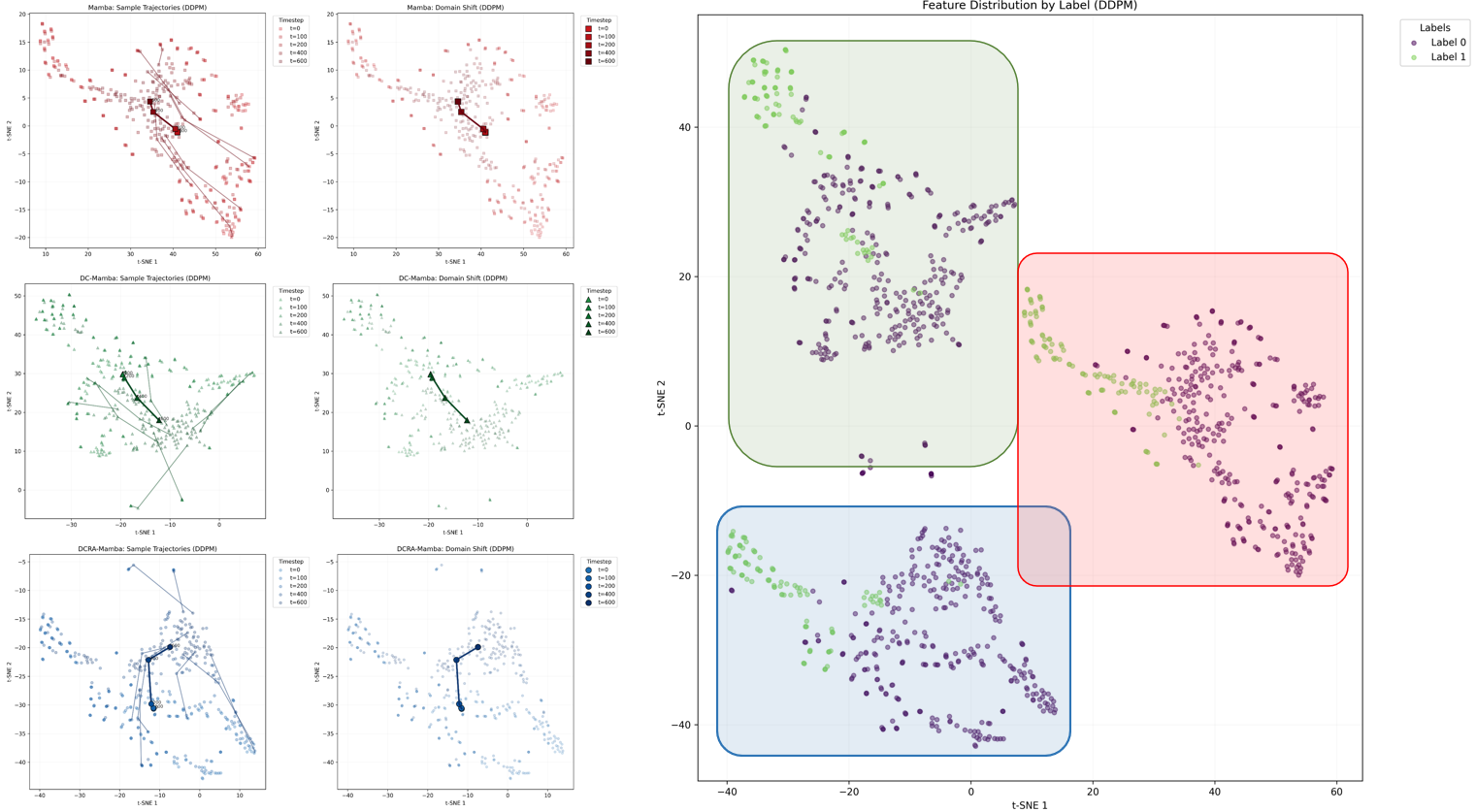}
        \caption{Feature Consistency Analysis of Mamba-based Models under Diffusion Noise}
        \label{fig:mamba_consistency}
    \end{subfigure}

    \caption{
    Visualization of feature evolution across diffusion timesteps for different encoders via t-SNE. For each backbone, the top, middle, and bottom rows correspond to the baseline, DC, and DCRA encoders, respectively. From left to right, we show sample-wise trajectories, domain shift behavior, and feature distributions colored by class labels. The full DCRA framework produces smoother and more coherent trajectories that follow the underlying data manifold, while preserving clearer class-dependent structures under increasing noise levels.
    }
    \label{fig:feature_consistency}
\end{figure*}

At the end, we examine the learned representations via visualization of feature distributions and trajectories across multiple noise levels as shown in Fig.~\ref{fig:feature_consistency}. This analysis provides insights into the impact of different training strategies on cross-noise consistency, trajectory smoothness, and class structure in the latent space. From Fig~\ref{fig:feature_consistency}, it is clear that the baseline encoders exhibit relatively stable but rigid feature evolution, with class imbalance in representation consistency. In contrast, the DC-encoders introduce stronger disturbances, which lead to irregular and unstable trajectories that deviate from the data manifold. Benefiting from the feature consistency mechanism, the DCRA-encoders enforce smoother trajectories, where features evolve by following coherent paths and remain aligned with the intrinsic geometry of the data. These observations are also supported by quantitative perspectives. The baseline models achieve the highest raw cosine similarity across timesteps, but they exhibit significant imbalance in intra-class consistency. 
DCRA maintains comparable cross-timestep similarity as well as improving class-balanced consistency. For instance, in Transformer-based models, DCRA improves the class cosine similarity for positive labels from $0.898$ to $0.945$, and in Mamba-based models from $0.862$ to $0.937$. At the same time, DCRA avoids the excessive feature drift observed in DC, which leads to shorter and more coherent trajectory lengths.

Diffusion conditioning with time embedding provides a strong mechanism for improving robustness under noise, while the addition of feature consistency further shapes the representation space into a more semantically meaningful structure. This leads to more stable and interpretable features, which may be more helpful for downstream tasks.

\section{Conclusion}



In this paper, we propose DCRA, a novel diffusion-conditioned framework for robust time-series learning. The proposed framework enhances encoders to learn representations that are robust to noise and distribution shifts by leveraging the forward diffusion process as a structured corruption scheduler. DCRA improves robustness under low-SNR and low-FPR conditions on the CHB-MIT dataset. 

We further introduce cross-noise consistency regularization to enforce smooth and class-discriminative feature evolution across noise levels. Experimental results show that DCRA produces more stable and semantically coherent representations than both baseline and diffusion-conditioned-only models. 
These findings suggest that structured diffusion corruption combined with representation alignment provides an effective strategy for robust time-series representation learning. Our findings highlight that diffusion conditioning and feature consistency play complementary roles within the DCRA framework for robust time-series representation learning. These noise-robust representations may provide a strong foundation for a wide range of downstream tasks, including subject-specific learning, cross-subject generalization, and multi-modal biomedical signal analysis. Future work will extend the proposed framework into broader encoder architectures and investigate its applicability to other time-series learning domains.

\bibliography{reference}

@misc{gu2020hipporecurrentmemoryoptimal,
      title={{HiPPO: Recurrent Memory with Optimal Polynomial Projections}}, 
      author={Albert Gu and Tri Dao and Stefano Ermon and Atri Rudra and Christopher Re},
      year={2020},
      eprint={2008.07669},
      archivePrefix={arXiv},
      primaryClass={cs.LG},
      url={https://arxiv.org/abs/2008.07669}, 
}

@misc{gu2024mambalineartimesequencemodeling,
      title={{Mamba: Linear-Time Sequence Modeling with Selective State Spaces}}, 
      author={Albert Gu and Tri Dao},
      year={2024},
      eprint={2312.00752},
      archivePrefix={arXiv},
      primaryClass={cs.LG},
      url={https://arxiv.org/abs/2312.00752}, 
}

@misc{sohldickstein2015deepunsupervisedlearningusing,
      title={{Deep Unsupervised Learning using Nonequilibrium Thermodynamics}}, 
      author={Jascha Sohl-Dickstein and Eric A. Weiss and Niru Maheswaranathan and Surya Ganguli},
      year={2015},
      eprint={1503.03585},
      archivePrefix={arXiv},
      primaryClass={cs.LG},
      url={https://arxiv.org/abs/1503.03585}, 
}

@misc{ho2020denoisingdiffusionprobabilisticmodels,
      title={{Denoising Diffusion Probabilistic Models}}, 
      author={Jonathan Ho and Ajay Jain and Pieter Abbeel},
      year={2020},
      eprint={2006.11239},
      archivePrefix={arXiv},
      primaryClass={cs.LG},
      url={https://arxiv.org/abs/2006.11239}, 
}

@InProceedings{Hendrycks_2021_ICCV,
    author    = {Hendrycks, Dan and Basart, Steven and Mu, Norman and Kadavath, Saurav and Wang, Frank and Dorundo, Evan and Desai, Rahul and Zhu, Tyler and Parajuli, Samyak and Guo, Mike and Song, Dawn and Steinhardt, Jacob and Gilmer, Justin},
    title     = {{The Many Faces of Robustness: A Critical Analysis of Out-of-Distribution Generalization}},
    booktitle = {Proceedings of the IEEE/CVF International Conference on Computer Vision (ICCV)},
    month     = {October},
    year      = {2021},
    pages     = {8340-8349}
}

@article{song2020score,
  title={{Score-based generative modeling through stochastic differential equations}},
  author={Song, Yang and Sohl-Dickstein, Jascha and Kingma, Diederik P and Kumar, Abhishek and Ermon, Stefano and Poole, Ben},
  journal={arXiv preprint arXiv:2011.13456},
  year={2020}
}

@article{cubuk1805autoaugment,
  title={{Autoaugment: Learning augmentation policies from data. arXiv 2018}},
  author={Cubuk, Ekin D and Zoph, Barret and Mane, Dandelion and Vasudevan, Vijay and Le, Quoc V},
  journal={arXiv preprint arXiv:1805.09501},
  volume={2},
  year={2018}
}

@inproceedings{cubuk2020randaugment,
  title={{Randaugment: Practical automated data augmentation with a reduced search space}},
  author={Cubuk, Ekin D and Zoph, Barret and Shlens, Jonathon and Le, Quoc V},
  booktitle={Proceedings of the IEEE/CVF conference on computer vision and pattern recognition workshops},
  pages={702--703},
  year={2020}
}

@article{zhang2017mixup,
  title={{mixup: Beyond empirical risk minimization}},
  author={Zhang, Hongyi and Cisse, Moustapha and Dauphin, Yann N and Lopez-Paz, David},
  journal={arXiv preprint arXiv:1710.09412},
  year={2017}
}

@inproceedings{chen2020simple,
  title={{A simple framework for contrastive learning of visual representations}},
  author={Chen, Ting and Kornblith, Simon and Norouzi, Mohammad and Hinton, Geoffrey},
  booktitle={International conference on machine learning},
  pages={1597--1607},
  year={2020},
  organization={PmLR}
}

@inproceedings{he2020momentum,
  title={{Momentum contrast for unsupervised visual representation learning}},
  author={He, Kaiming and Fan, Haoqi and Wu, Yuxin and Xie, Saining and Girshick, Ross},
  booktitle={Proceedings of the IEEE/CVF conference on computer vision and pattern recognition},
  pages={9729--9738},
  year={2020}
}

@article{tarvainen2017mean,
  title={{Mean teachers are better role models: Weight-averaged consistency targets improve semi-supervised deep learning results}},
  author={Tarvainen, Antti and Valpola, Harri},
  journal={Advances in neural information processing systems},
  volume={30},
  year={2017}
}

@article{sohn2020fixmatch,
  title={{Fixmatch: Simplifying semi-supervised learning with consistency and confidence}},
  author={Sohn, Kihyuk and Berthelot, David and Carlini, Nicholas and Zhang, Zizhao and Zhang, Han and Raffel, Colin A and Cubuk, Ekin Dogus and Kurakin, Alexey and Li, Chun-Liang},
  journal={Advances in neural information processing systems},
  volume={33},
  pages={596--608},
  year={2020}
}

@inproceedings{mittal2023diffusion,
  title={{Diffusion Based Representation Learning}},
  author={Mittal, Sarthak and Abstreiter, Korbinian and Bauer, Stefan and Sch{\"o}lkopf, Bernhard and Mehrjou, Arash},
  booktitle={International conference on machine learning},
  pages={24963--24982},
  year={2023},
  organization={PMLR}
}

@inproceedings{xie2020self,
  title={Self-training with noisy student improves imagenet classification},
  author={Xie, Qizhe and Luong, Minh-Thang and Hovy, Eduard and Le, Quoc V},
  booktitle={Proceedings of the IEEE/CVF conference on computer vision and pattern recognition},
  pages={10687--10698},
  year={2020}
}

@article{oliver2018realistic,
  title={Realistic evaluation of deep semi-supervised learning algorithms},
  author={Oliver, Avital and Odena, Augustus and Raffel, Colin A and Cubuk, Ekin Dogus and Goodfellow, Ian},
  journal={Advances in neural information processing systems},
  volume={31},
  year={2018}
}

@article{Deng_2022,
   title={ArcFace: Additive Angular Margin Loss for Deep Face Recognition},
   volume={44},
   ISSN={1939-3539},
   url={http://dx.doi.org/10.1109/TPAMI.2021.3087709},
   DOI={10.1109/tpami.2021.3087709},
   number={10},
   journal={IEEE Transactions on Pattern Analysis and Machine Intelligence},
   publisher={Institute of Electrical and Electronics Engineers (IEEE)},
   author={Deng, Jiankang and Guo, Jia and Yang, Jing and Xue, Niannan and Kotsia, Irene and Zafeiriou, Stefanos},
   year={2022},
   month=Oct, pages={5962–5979} }

@inproceedings{wen2016discriminative,
  title={A discriminative feature learning approach for deep face recognition},
  author={Wen, Yandong and Zhang, Kaipeng and Li, Zhifeng and Qiao, Yu},
  booktitle={European conference on computer vision},
  pages={499--515},
  year={2016},
  organization={Springer}
}

@article{snell2017prototypical,
  title={Prototypical networks for few-shot learning},
  author={Snell, Jake and Swersky, Kevin and Zemel, Richard},
  journal={Advances in neural information processing systems},
  volume={30},
  year={2017}
}

@misc{qiang2024ecgmambaefficientecgclassification,
      title={{ECGMamba: Towards Efficient ECG Classification with BiSSM}}, 
      author={Yupeng Qiang and Xunde Dong and Xiuling Liu and Yang Yang and Yihai Fang and Jianhong Dou},
      year={2024},
      eprint={2406.10098},
      archivePrefix={arXiv},
      primaryClass={cs.LG},
      url={https://arxiv.org/abs/2406.10098}, 
}

@misc{puah2025eegdmeegrepresentationlearning,
      title={{EEGDM: EEG Representation Learning via Generative Diffusion Model}}, 
      author={Jia Hong Puah and Sim Kuan Goh and Ziwei Zhang and Zixuan Ye and Chow Khuen Chan and Kheng Seang Lim and Si Lei Fong and Kok Sin Woon and Cuntai Guan},
      year={2025},
      eprint={2508.14086},
      archivePrefix={arXiv},
      primaryClass={cs.LG},
      url={https://arxiv.org/abs/2508.14086}, 
}

@article{islam2016methods,
  title={{Methods for artifact detection and removal from scalp EEG: A review}},
  author={Islam, Md Kafiul and Rastegarnia, Amir and Yang, Zhi},
  journal={Neurophysiologie Clinique/Clinical Neurophysiology},
  volume={46},
  number={4-5},
  pages={287--305},
  year={2016},
  publisher={Elsevier}
}

@article{obeid2016temple,
  title={{The temple university hospital EEG data corpus}},
  author={Obeid, Iyad and Picone, Joseph},
  journal={Frontiers in neuroscience},
  volume={10},
  pages={196},
  year={2016},
  publisher={Frontiers Media SA}
}

@article{PhysioNet-chbmit-1.0.0,
  author = {Guttag, John},
  title = {{CHB-MIT Scalp EEG Database}},
  journal = {{PhysioNet}},
  year = {2010},
  month = jun,
  note = {Version 1.0.0},
  doi = {10.13026/C2K01R},
  url = {https://doi.org/10.13026/C2K01R}
}

@article{lawhern2018eegnet,
  title={{EEGNet: a compact convolutional neural network for EEG-based brain--computer interfaces}},
  author={Lawhern, Vernon J and Solon, Amelia J and Waytowich, Nicholas R and Gordon, Stephen M and Hung, Chou P and Lance, Brent J},
  journal={Journal of neural engineering},
  volume={15},
  number={5},
  pages={056013},
  year={2018},
  publisher={iOP Publishing}
}

@article{bashivan2015learning,
  title={{Learning representations from EEG with deep recurrent-convolutional neural networks}},
  author={Bashivan, Pouya and Rish, Irina and Yeasin, Mohammed and Codella, Noel},
  journal={arXiv preprint arXiv:1511.06448},
  year={2015}
}

@article{song2022eeg,
  title={{EEG conformer: Convolutional transformer for EEG decoding and visualization}},
  author={Song, Yonghao and Zheng, Qingqing and Liu, Bingchuan and Gao, Xiaorong},
  journal={IEEE Transactions on Neural Systems and Rehabilitation Engineering},
  volume={31},
  pages={710--719},
  year={2022},
  publisher={IEEE}
}

@article{zhang2021survey,
  title={A survey on deep learning-based non-invasive brain signals: recent advances and new frontiers},
  author={Zhang, Xiang and Yao, Lina and Wang, Xianzhi and Monaghan, Jessica and Mcalpine, David and Zhang, Yu},
  journal={Journal of neural engineering},
  volume={18},
  number={3},
  pages={031002},
  year={2021},
  publisher={IOP Publishing}
}

\end{document}